\documentclass[11pt,a4paper]{article}

\usepackage[T1]{fontenc}
\usepackage[utf8]{inputenc}
\usepackage{lmodern}
\usepackage{microtype}
\usepackage[a4paper,margin=1.3in]{geometry}

\usepackage{amsmath,amssymb}
\usepackage[hidelinks]{hyperref}
\usepackage{enumitem}

\newenvironment{keywords}{%
  \vspace{0.6em}\noindent\textbf{Keywords:}\ }%
  {\par\vspace{0.6em}}

\title{\textbf{Comment on arXiv:2511.21731v1:\\
``Identifying Quantum Structure in AI Language: Evidence for Evolutionary Convergence of Human and Artificial Cognition''}}
\author{Krzysztof Sienicki\thanks{Chair of Theoretical Physics of Naturally Intelligent Systems (\(\mathbb{NIS}\)\textsuperscript{\copyright}), Lipowa~2/Topolowa~19, 05–807 Podkowa Leśna, Poland, European Union; E-mail: \texttt{niskrissienicki@gmail.com}}%
}
\date{\today}

\begin{document}
\maketitle

\begin{abstract}
This note is a friendly technical check of arXiv:2511.21731v1. I highlight a few places where the manuscript’s
interpretation of (i) the reported CHSH/Bell-type calculations and (ii) Bose--Einstein (BE) fits to rank--frequency data
seems to go beyond what the stated procedures can firmly support. I also point out one internal inconsistency in the
``energy-level spacing'' analogy. The aim is constructive: to keep the interesting empirical observations, while making
clear what they do (and do not) imply about quantum entanglement in the usual Hilbert-space sense, especially when
``energy'' is defined by rank.
\end{abstract}


\section{Scope}
The preprint arXiv:2511.21731v1 develops two connected themes:
\begin{enumerate}[label=(\alph*),leftmargin=1.2em]
\item CHSH values computed from human questionnaire data and from responses of two large language models (LLMs);
\item fits of rank--frequency distributions of LLM-generated words by a Bose--Einstein (BE) occupancy form, interpreted
as evidence for ``quantum structure'' in AI language.
\end{enumerate}

Both themes are interesting as \emph{phenomenology} and as links to the quantum-cognition literature. My concern is mainly
about interpretation: a few statements in the current version read stronger than what follows from the reported protocol
unless one adds extra operational controls or adopts a more careful framing.

\section{Main points}

\subsection{A CHSH computation on prompts is not, by itself, a Bell test}
A Bell/CHSH test is more than ``evaluate a CHSH expression under four conditions.'' The usual Bell inference relies on a
specific operational structure and, formally, on assumptions such as locality/factorization and measurement independence.
One common way to state this is to posit hidden variables $\lambda$ such that
\begin{equation}
P(a,b\mid x,y,\lambda)=P(a\mid x,\lambda)\,P(b\mid y,\lambda)
\label{eq:factorization}
\end{equation}
(local causality / factorization) together with
\begin{equation}
P(\lambda\mid x,y)=P(\lambda)
\label{eq:measindep}
\end{equation}
(measurement independence / ``freedom of choice'').
In actual experiments, considerable effort goes into making \eqref{eq:measindep} as plausible as possible (for example,
the freedom-of-choice emphasis in the Big Bell Test Collaboration \cite{Abellan2018}).

In the LLM procedure described in the manuscript, the ``settings'' are embedded in the prompts given to a text generator
(or, in the relevant table, the reported probabilities are averaged across two different models). In this situation the
model’s internal state is typically \emph{prompt-dependent}, so the setting-independence logic behind \eqref{eq:measindep}
is not obviously in place. For that reason, a CHSH value computed from these responses should not be read as evidence of
nonlocality or physical entanglement. A safer takeaway is:
\emph{the conditional response statistics are strongly context-dependent and may not admit a single global, noncontextual
(Kolmogorovian) representation}.

If the authors want a more Bell-like operational claim, the protocol would need to track the Bell structure more closely:
isolated instances playing ``Alice'' and ``Bob,'' truly exogenous randomized settings, repeated trials per setting under
controlled sampling conditions (temperature/seed/new-chat policy), and uncertainty estimates.

\subsection{CHSH = 4 is supra-quantum, not ``stronger quantum entanglement''}
Using one common convention, the CHSH expression is
\begin{equation}
S = E(A_0B_0)+E(A_0B_1)+E(A_1B_0)-E(A_1B_1),
\end{equation}
with the local-hidden-variable bound $|S|\le 2$ \cite{CHSH1969}. Quantum theory satisfies Tsirelson's bound
$|S|\le 2\sqrt{2}$ \cite{Tsirelson1980}. The algebraic maximum is $|S|=4$ (a PR box is the canonical
\emph{nonsignalling} extremal that attains this value).

The manuscript reports an LLM-derived value $S=4$ and describes this as ``much stronger entanglement.'' If ``quantum'' is
meant in the standard Hilbert-space sense, that interpretation does not hold: $S=4$ exceeds Tsirelson's bound and is
therefore \emph{supra-quantum}. A consistent reading would be either:
(i) treat the LLM result as showing extremely strong contextual constraints in completions, or
(ii) explicitly adopt a generalized-probabilistic-theory (GPT) framing and avoid language suggesting
``stronger quantum entanglement.''

\subsection{Inference for the human CHSH violation: the stated t-test needs justification}
For the human data, the manuscript reports a CHSH value above $2$ and quotes a p-value from a one-sample t-test against
$2$. Since CHSH is computed from aggregated response frequencies across several question-conditions, it is not obvious
what the t-test’s sampling unit is, or why normal/iid assumptions would apply in the stated form.

More standard options would be:
\begin{itemize}[leftmargin=1.2em]
\item compute participant-level CHSH values (if the protocol supports per-participant quadruples) and analyze those;
\item use a bootstrap/permutation analysis on participant responses or on the multinomial counts to obtain a confidence
interval for $S$ directly.
\end{itemize}

\subsection{Internal inconsistency in the ``particle in a box'' spacing exponent}
The manuscript maps ``rank $\rightarrow$ energy level'' and introduces an exponent $d$ so that energy spacings scale like
$n^d$. It correctly notes (in words) that the textbook ``particle in a box'' has energies growing like $n^2$ (so spacings
widen with $n$), but later assigns $d=-2$ to the particle-in-a-box case. This looks like a simple sign/value slip: for the
standard particle in a box one expects $d=+2$.

Fixing this would make the analogy section internally consistent.

\subsection{BE fits to rank--frequency data: what they can (and cannot) diagnose under rank-as-energy}
The manuscript presents BE-versus-MB curve fits as evidence for ``quantum structure within the code.'' Here it helps to
keep a clear distinction between \emph{a useful phenomenological fit} and \emph{evidence for a specific physical
mechanism}. Related cautions have been discussed in the BE-in-language context; for convenience, I summarize the most
relevant ones here (see \cite{Sienicki2025BEComment} for a fuller discussion).

\paragraph{(i) Normalization is not a probability ``boost.''}
If an argument for ``bosonic enhancement'' hinges on the appearance of a factor $\sqrt{2}$ (or $2$ in the norm) when two
single-particle states are set equal inside an (un)normalized symmetrized expression, that factor by itself is not a
probability. While (anti)symmetrization changes the physical state for identical particles, an \emph{overall norm factor}
in an unnormalized expression is not itself a probability. Probabilities are computed from \emph{normalized} states; an
overall rescaling of a ket does not change observable statistics. Genuine bosonic bunching (e.g.\ Hong--Ou--Mandel
interference) appears only once a concrete measurement scenario and observable are specified \cite{HongOuMandel1987}. It
is therefore best to avoid reading overall norms of unnormalized expressions as propensities.

\paragraph{(ii) Rank-as-energy makes the BE fit only weakly diagnostic.}
The manuscript identifies ``energy levels'' with rank. If one sets
\begin{equation}
E_i = i,
\end{equation}
then a BE-shaped occupancy written in rank form can be parameterized as
\begin{equation}
N(i) = \frac{1}{A e^{i/B}-1}.
\label{eq:BE_rank}
\end{equation}
This is not an empirically measured constraint in the statistical-mechanics sense; it is a definition attached to the
rank table. So even an excellent BE-shaped fit does not, by itself, identify a ``bosonic'' generative mechanism unless
the mapping $E_i=i$ is operationally justified and shown robust under reasonable alternatives.

With this rank-as-energy choice, \eqref{eq:BE_rank} can also reproduce familiar linguistic scalings over finite ranges
\emph{when the fitted parameters place the data in an $i/B\ll 1$ regime across the ranks of interest}. For $i\ll B$ one
has $e^{i/B}=1+i/B+O((i/B)^2)$, hence
\begin{equation}
N(i)\approx \frac{1}{(A-1)+(A/B)i}.
\label{eq:BE_small_i}
\end{equation}
If $A$ is close to $1$, \eqref{eq:BE_small_i} is approximately Zipf--Mandelbrot-like, and in a window where the offset
term is negligible it becomes approximately Zipf-like ($N(i)\propto 1/i$) \cite{Zipf1935,Mandelbrot1953}. This does not
undermine the fit; it simply shows why, under $E_i=i$, a BE-shaped curve is \emph{not uniquely diagnostic} of a quantum
mechanism.

For sufficiently large $i$ with $A e^{i/B}\gg 1$, one also has the exponential-tail approximation
\begin{equation}
N(i)\approx \frac{1}{A}\,e^{-i/B},
\end{equation}
so a soft high-rank decay is expected in this parametrization.

\paragraph{(iii) The MB-labelled baseline is too weak for an ontological conclusion.}
The manuscript contrasts BE fits with an exponential form labelled ``Maxwell--Boltzmann.'' Under rank-as-energy, that
MB-labelled baseline is essentially an exponential-in-rank comparator, which is not a strong competitor for heavy-tailed
rank--frequency data. Therefore, ``BE fits better than MB'' is not, by itself, evidence that the underlying mechanism is
bosonic or ``quantum.''

A stronger and more standard approach would compare BE-shaped fits against common linguistic/statistical baselines
(Zipf--Mandelbrot, log-normal, stretched exponential, Pitman--Yor/Simon-type models), using discrete likelihoods
appropriate for counts, uncertainty estimates, and out-of-sample validation. Likelihood-based criteria such as AIC/BIC
can be reported where appropriate \cite{ClausetShaliziNewman2009}.

\paragraph{(iv) Parameter semantics and reproducibility.}
If $B$ is interpreted literally as a ``temperature'' (or a close thermodynamic analogue), that interpretation calls for an
ensemble derivation with clearly stated constraints and a justified density of states, not just a fit in the variable $i$
\cite{PathriaBealeSM}. Finally, quantitative claims should state the corpus choice, preprocessing steps (tokenization,
case-folding, lemmatization/stopwords), fitting procedure, uncertainty estimates, and the model-selection criterion.

\section{Concluding suggestion}
In my view, the paper would be stronger if the CHSH section were framed explicitly as evidence of \emph{contextuality in
responses} (unless a Bell-like trial protocol is implemented), if the spacing-exponent slip were corrected, and if the BE
fits were presented as phenomenology supported by comparisons to standard linguistic baselines. These changes would keep
the manuscript's central ideas intact while aligning the conclusions more closely with what the reported procedures can
genuinely support.


\vspace{0.5em}
\begin{thebibliography}{99}

\bibitem{Bell1964}
J.~S.~Bell, ``On the Einstein Podolsky Rosen paradox,'' \emph{Physics} \textbf{1}, 195--200 (1964).

\bibitem{CHSH1969}
J.~F.~Clauser, M.~A.~Horne, A.~Shimony, and R.~A.~Holt,
``Proposed experiment to test local hidden-variable theories,''
\emph{Phys.\ Rev.\ Lett.} \textbf{23}, 880--884 (1969).
\href{https://doi.org/10.1103/PhysRevLett.23.880}{doi:10.1103/PhysRevLett.23.880}

\bibitem{Tsirelson1980}
B.~S.~Tsirelson, ``Quantum generalizations of Bell's inequality,''
\emph{Lett.\ Math.\ Phys.} \textbf{4}, 93--100 (1980).
\href{https://doi.org/10.1007/BF00417500}{doi:10.1007/BF00417500}

\bibitem{Abellan2018}
C.~Abell\'an \emph{et al.} (The BIG Bell Test Collaboration),
``Challenging local realism with human choices,''
\emph{Nature} \textbf{557}, 212--216 (2018).
\href{https://doi.org/10.1038/s41586-018-0085-3}{doi:10.1038/s41586-018-0085-3}

\bibitem{HongOuMandel1987}
C.~K.~Hong, Z.~Y.~Ou, and L.~Mandel,
``Measurement of subpicosecond time intervals between two photons by interference,''
\emph{Phys.\ Rev.\ Lett.} \textbf{59}, 2044--2046 (1987).
\href{https://doi.org/10.1103/PhysRevLett.59.2044}{doi:10.1103/PhysRevLett.59.2044}

\bibitem{Zipf1935}
G.~K.~Zipf,
\emph{The Psycho-Biology of Language: An Introduction to Dynamic Philology},
Houghton Mifflin (1935).

\bibitem{Mandelbrot1953}
B.~Mandelbrot,
``An informational theory of the statistical structure of language,''
in \emph{Communication Theory}, ed. W.~Jackson, Butterworths (1953).

\bibitem{ClausetShaliziNewman2009}
A.~Clauset, C.~R.~Shalizi, and M.~E.~J.~Newman,
``Power-law distributions in empirical data,''
\emph{SIAM Review} \textbf{51}(4), 661--703 (2009).
\href{https://doi.org/10.1137/070710111}{doi:10.1137/070710111}

\bibitem{PathriaBealeSM}
R.~K.~Pathria and P.~D.~Beale,
\emph{Statistical Mechanics}, 3rd ed., Academic Press (2011).

\bibitem{Sienicki2025BEComment}
M.~Sienicki and K.~Sienicki,
``Revised comment on `The Origin of Quantum Mechanical Statistics: Insights from Research on Human Language'
(arXiv:2407.14924),'' arXiv:2512.07881v1 (2025).

\bibitem{Target}
S.~Song and C.-H.~Choi,
``Identifying Quantum Structure in AI Language: Evidence for Evolutionary Convergence of Human and Artificial Cognition,''
arXiv:2511.21731v1 (2025).
\href{https://arxiv.org/abs/2511.21731}{arXiv:2511.21731}

\end{thebibliography}
\end{document}